\documentclass[graybox]{svmult}
\usepackage{geometry}\usepackage[justification=centering]{caption}
\usepackage{amsmath}
\usepackage[table, dvipsnames]{xcolor}
\usepackage{amssymb}

\usepackage{mathptmx}       % selects Times Roman as basic font
\usepackage{helvet}         % selects Helvetica as sans-serif font
\usepackage{courier}        % selects Courier as typewriter font
\usepackage{type1cm}        % activate if the above 3 fonts are
\usepackage{makeidx}         % allows index generation
\usepackage{graphicx}        % standard LaTeX graphics tool
\usepackage{tabularx}
\usepackage{float}
\usepackage{subcaption}
\usepackage{multirow}
\usepackage{textcomp}
\usepackage{multicol}        % used for the two-column index
\usepackage[bottom]{footmisc}% places footnotes at page bottom

\makeindex             % used for the subject index
\begin{document}

\title*{From Zero-Shot Machine Learning to Zero-Day Attack Detection}
% \title*{Zero-Day Attack Detection in a Zero-Shot Learning Setup}
\author{Mohanad Sarhan, Siamak Layeghy, Marcus Gallagher and Marius Portmann}
\institute{Mohanad Sarhan \at The University of Queensland, Brisbane QLD 4072, \email{m.sarhan@uq.net.au}
\and Siamak Layeghy \at The University of Queensland, Brisbane QLD 4072, \email{siamak.layeghy@uq.net.au}
\and Marius Portmann \at The University of Queensland, Brisbane QLD 4072, \email{marius@itee.uq.edu.au}}
%
% Use the package "URL.sty" to avoid
\maketitle

\abstract{Machine Learning (ML) models have been proven to be efficient in the classification and prediction of test data samples to their respective categories. The standard ML methodology assumes that the test samples are derived from a set of pre-observed classes used in the training phase. Where the model extracts and learns useful patterns to detect new data samples belonging to the same data classes. However, in certain applications such as Network Intrusion Detection Systems (NIDSs), it is challenging to obtain data samples for all attack classes that the model will most likely observe in production. ML-based NIDSs face new attack traffic known as zero-day attacks, that are not used in the training of the learning models due to their non-existence at the time. Therefore, in this paper, a zero-shot learning methodology has been proposed to evaluate the ML model performance in the detection of zero-day attack scenarios. In the attribute learning stage, the ML models map the network data features to distinguish semantic attributes from known attack (seen) classes. In the inference stage, the models are evaluated in the detection of zero-day attack (unseen) classes by constructing the relationships between known attacks and zero-day attacks. A new metric is defined as \textit{Zero-day Detection Rate}, which measures the effectiveness of the learning model in the inference stage. The results demonstrate that while the majority of the attack classes do not represent significant risks to organisations adopting an ML-based NIDS in a zero-day attack scenario. However, for certain attack groups identified in this paper, such systems are not effective in applying the learnt attributes of attack behaviour to detect them as malicious. Further Analysis was conducted using the Wasserstein Distance technique to measure how different such attacks are from other attack types used in the training of the ML model. The results demonstrate that sophisticated attacks with a low zero-day detection rate have a significantly distinct feature distribution compared to the other attack classes.}

\keywords{
Machine Learning, Network Intrusion Detection System, Zero-day attack, Zero-shot learning
}

\section{Introduction}
%%ML%%
Over the past few years, Machine Learning (ML) capabilities have been utilised to enhance the performance and efficiency of various technological applications \cite{ghahramani2015probabilistic}. ML is a subset of Artificial Intelligence (AI) \cite{panch2018artificial}, involving a set of statistical algorithms that can learn from data without being explicitly programmed \cite{Koza1996}. ML models are recognised for their superior ability to extract and learn complex data patterns that are not feasibly realizable by domain experts \cite{najafabadi2015deep}. The learnt patterns are used to predict, classify, and regress future events and scenarios. This has been a disruptive innovation  \cite{bloomfield2019disruptive} in multiple industries where operational automation and efficiency are required. Therefore, ML models have been widely deployed across multiple domains, proving great success over traditional computing algorithms, where it is challenging to perform the required operations. The same motivation has led to the implementation of ML models in the cybersecurity domain \cite{buczak2015survey}, to further enhance and strengthen the security posture of organisations. The intelligence of ML models has been taken advantage of in securing computer networks \cite{apruzzese2018effectiveness} against advanced threats. The addition of the intelligence element to the organisation's security strategy will add sophisticated layers of defence \cite{dua2016data} that can limit the number of internal and external threats if designed efficiently. ML operation is capable of detecting complicated modern attacks that require advanced innovation detection capabilities \cite{alrashdi2019ad}.

%%NIDS%%
Network Intrusion Detection Systems (NIDSs) are essential security tools that detect threats as they penetrate an organisation's network environment \cite{mukherjee1994network}. There are two main types of NIDS; Signature-based NIDSs scan incoming network traffic for any Indicator of Compromise (IOC), also known as attack signatures, such as source IPs, domain names, and hash values, that might indicate malicious traffic \cite{kumar2012signature}. One of the main and ongoing challenges of securing computer networks with signature-based NIDSs is the detection of zero-day attacks \cite{garcia2009anomaly}. A Zero-day attack is a new kind of threat that has not been seen before \cite{bilge2012before}, designed to infiltrate or disrupt network communications. It is an unknown vulnerability to security administrators, that hackers can exploit before it has been remediated. A recent example is a zero-day vulnerability discovered in Microsoft Windows in June 2019, that was targeting local escalation privileges \cite{stellios2019advanced}. Generally, when a zero-day attack is discovered, it gets added to the publicly shared Common Vulnerabilities and Exposures (CVE) list \cite{mell2002use}. Known security vulnerabilities are defined using a CVE code and severity level \cite{mell2002use}, and shared amongst the community for immediate action. The remediation of zero-day attacks is generally conducted by adding IOCs related to the threat into a list of detection databases \cite{ganame2017network} used by signature-based NIDS. As such, signature-based NIDSs are deemed to be unreliable in the detection of zero-day attacks simply because the complete set of IOCs has not been discovered or registered for monitoring at the time of penetration. 

%%ML-NIDS%%

It is very challenging to identify zero-day attacks using signature-based NIDS, as it takes an average of 312 days \cite{bilge2012before} to obtain the full set of attack IOCs. Meanwhile, organisations protected by signature-based NIDS are vulnerable to such attacks. Therefore, the focus has diverted towards building ML-based NIDS \cite{garcia2009anomaly}, an enhanced modern edition of traditional NIDS to overcome the limitations faced in the detection of zero-day or unseen attacks. ML-based NIDS are designed and deployed to scan and analyse incoming network traffic for any anomalies or malicious intent \cite{garcia2009anomaly}. The Analysis process is conducted via a comparison of the incoming network behaviour with the learnt behaviour of safe and intrusive traffic \cite{sinclair1999application}. During the design process, the ML model is trained using a set of benign and attack samples, where the hidden complex pattern of traffic is learned. Unlike the signature-based NIDSs that solely relies on IOC for detection, the ML-based NIDS utilises the learnt behavioural pattern to detect network attacks \cite{sinclair1999application}. This has a great potential of detecting zero-day attacks as the requirement of obtaining IOC is obsolete \cite{sahu2015network}. Zero-day attacks can be detected by ML-based NIDS using the learned attack behaviour, which attracts great attention and focus towards the development of such models. Most of the available research works have aimed at the design and evaluation of ML-based NIDSs in the detection of known attack groups. However, a limited amount of research has focused on the evaluation of zero-day attack detection to measure the benefits of ML-based NIDS over signature-based NIDS. As such, a large number of proposed ML-based NIDS does not consider the most likely re-occurring scenario of zero-day attacks, where a new attack class may appear after the learning stage of the ML model.

%%zeroshot
Zero-shot Learning (ZSL) is an emerging methodology used to evaluate and improve the generalisability of ML models to new or unseen data classes \cite{xian2017zero}.  This technique follows the assumption that the training dataset might not include the entire set of classes that the ML model could observe once deployed in the real world. As such, ZSL addresses the ever-growing set of classes that might render it unfeasible to collect training samples for each of them \cite{wang2019survey}. ZSL involves the recognition of new data samples derived from previously unseen classes. In the attribute learning stage, the model is provided with distinguishing semantics of the missing class. In the inference stage, ZSL applies the learnt attributes to predict or classify certain samples belonging to the missing data class \cite{xian2017zero}. ZSL addresses one of the main challenges in building a reliable ML-based NIDS, which is the evaluation of detecting new attack classes that are not available in the training phase, such as zero-day attacks \cite{zhang2020unknown}. This directly applies to ML-based NIDS, as new attack classes that the learning model did not train on, are emerged and observed in the real world post-deployment. This includes zero-day attacks which could lead to fatal consequences to the adopting organisation if undetected \cite{bilge2012before}. Therefore, a reliable ML-based NIDS needs not only to be evaluated across a set of known attacks but also unknown attacks that were missing from the training dataset, simulating the likely scenario of a zero-day threat.

%%This paper experiments and contributions%%
In this paper, a new zero-day evaluation methodology, inspired by ZSL, is proposed. The framework measures how well an ML-based NIDS can detect unseen attacks using a set of semantic attributes learnt from seen attacks. There are two main stages of the proposed setup. In the attribute learning stage, the models extract and map the network data features to the unique attributes of known attacks. In the inference phase, the model associates the relationships between known attacks and zero-day attacks to assist in their discovery and classification as malicious. Unlike, traditional evaluation methods, the proposed setup aims to evaluate ML-based NIDS using a new metric, named Zero-day Detection Rate (Z-DR), that measures how well a learning model can reconstruct the distinguishing semantics learnt from known attack classes, to detect unknown attack classes. The proposed methodology has been implemented using a combination of two key NIDS datasets, each consisting of a broad range of modern attacks, and two widely used ML models in the research field. Additionally, the achieved results have been analysed using the Wasserstein Distance to explain the variation of the Z-DR with different attack groups. The key contribution of this paper is the adoption of a ZSL-based problem set up to propose a reliable evaluation methodology of ML-based NIDS in the detection of new or unseen attack types, mimicking the most likely occurrences of zero-day attacks post-deployment. In Section \ref{rw} key related works are discussed, followed by a detailed explanation of the proposed ZSL-based methodology in Section \ref{zsl}. The experimental methodology followed in this paper and the results obtained are discussed in Sections \ref{method} and \ref{results} respectively, before concluding this paper.

\section{Related Works}
\label{rw}
In this section, key related papers that aimed to evaluate NIDS for the detection of zero-day attacks are discussed. While most papers aimed to design sophisticated ML-based NIDS \cite{sommer2010outside}, the focus has been towards the evaluation of the proposed systems across a range of known attacks. Where traditional signature-based NIDSs have been achieving a satisfactory performance throughout the years in the detection of known attacks. Therefore, it is surprising to notice that only a few papers have aimed to challenge ML-based NIDS in the detection of unknown or zero-day attacks. In the case of unsupervised anomaly detection systems, where the model only learns the behaviour of benign traffic, the NIDS fundamentally work to detect each attack type as an unknown attack. However, it is noted that such models lead to a large number of false alarms leading to alert fatigue \cite{casas2012unsupervised}, as it does not consider the attack behaviour. Overall, there is a limited number of papers following a zero-shot learning methodology to detect zero-day attacks. Out of these works, none, to the best of our knowledge, have aimed to utilise modern network datasets which represent current network traffic characteristics, to evaluate their approach.

In \cite{holm2014signature}, the author has evaluated the zero-day attack detection performance using a signature-based NIDS. The paper studies the frequent claim that such systems are not capable of detecting zero-day attacks. The experiment studies 356 network attacks, out of which 183 attacks are unknown (zero-day) to the rule set. The paper utilised the Snort tool, a well-known signature-based NIDS in the industry. The Metasploit Framework is utilized to simulate the attack scenarios. The detection rate is calculated by applying a Snort ruleset which does not disclose the vulnerabilities relevant to the attack. The results show that Snort has an unreliable detection rate of 17\% against zero-day attacks. The paper argues that the frequent claim that signature-based NIDSs are not capable of detection of zero-day attacks is incorrect, as 17\% is significantly larger than zero. The author mentions that further mechanisms should be implemented to complement signature-based NIDS in the detection of unregistered attacks and the results of this paper can be seen as a baseline for zero-day attack detection.

In \cite{hindy2020utilising}, Hindy et al. aimed to improve the unsupervised outlier-based detection systems that usually suffer from high False Alarm Rate (FAR). The paper explored an autoencoder for the detection of zero-day attacks to maintain a high detection rate while lowering the FAR. The system is evaluated across two key datasets; CICIDS2017 and NSL-KDD. The methodology involved training the classifiers using the benign data samples and evaluating the detection of zero-day attacks. The results are compared to a one-class support vector machine, where the autoencoder is superior. The results demonstrate a zero-day detection accuracy of 89–99\% for the NSL-KDD dataset and 75–98\% for the
CICIDS2017 dataset. However, the proposed models do not take the attack behaviour into consideration and the numbers of undetected attacks and false alarms are unmeasured.

Zhang et al. \cite{zhang2020unknown}, has evaluated ML-based NIDS detection performance against zero-day attacks. The authors have used zero-shot learning to simulate the occurrence of zero-day attack scenarios. The ML models learn the distinguishing information between the attack and benign classes by mapping the feature space and attribute space. The authors utilised a sparse autoencoder model that projects the features of known attacks to a semantic space and establishes a feature to semantic mapping to detect the unknown attacks. The paper utilised the attacks present in the NSL-KDD dataset, released in 1998, to simulate a zero-day scenario, the dataset contains 4 attack scenarios. The results demonstrate that the average accuracy achieved is 88.3\% across the available attacks in the dataset.

Li et al. \cite{li2019zero} focused on attribute learning methods to detect unknown attack types. The authors followed a zero-shot learning method to design a NIDS to overcome the anomaly detection limitation faced by current methods. The architecture involves a pipeline using a Random Forest (RF) feature selection and a spatial clustering attribute conversion method. The results demonstrate that the proposed method overcomes the state-of-the-art approaches in anomaly detection. The attribute learning framework converts network data samples into unsupervised cluster attributes. The NSL-KDD dataset has been utilised to evaluate the proposed framework where it was able to detect the DoS (apache2) and Probe (saint) attacks achieving an overall accuracy of 34.71\%. The authors compared its performance with a decision tree classifier which achieved a poor overall accuracy of 13.59\%.

Overall, significant contributions have been provided by the research works aiming to evaluate the performance of ML-based NIDS in the detection of unknown attacks. However, only a very small number adopted a ZSL-based setup to simulate the occurrence of zero-day attacks.
% , which is challenging to reproduce or customize using alternative ML models due to the dependency of a function performed by a model such as sparse autoencoder in \cite{zhang2020unknown} and spatial clustering in \cite{li2019zero}. 
Moreover, there has been a very limited amount of experimental work on modern zero-day attack scenarios with recent data sets and attack types, which limits the identification of sophisticated attacks that can not be detected in zero-day scenarios. In addition, it is surprising that some recent works still utilise the NSL-KDD dataset for evaluation purposes, given that it is more than 20 years old. 
The attack scenarios available in the dataset do not represent modern network traffic characteristics and threats, which limits the reliability of the proposed methodology and its evaluation \cite{8672520}.

\section{Proposed Methodology}
\label{zsl}

%%ML

In a traditional ML evaluation methodology, the learning model is trained and tested on the same set of data classes. In the training stage, the model learns to identify patterns directly from each data class. In the testing stage, the model applies the learnt patterns to identify the data samples derived from the same data classes used in the training stage. In an experimental setup, the utilised dataset is split into training and testing partitions, where both sets have the same number and type of classes. The learning model is trained on the training set using the complete set of classes that also forms the test set used in the evaluation stage. This approach of evaluation follows the assumption that the dataset collected for the training of ML models includes the full set of classes that the model will observe post-deployment in production. In the case of an ML-based NIDS, the model is trained and tested using a set of known attack classes. The model is evaluated on how well it can detect data samples derived from known attack groups as malicious.

% The training set $D_{tr}$, and testing set $D_{tst}$ of an NIDS dataset can be represented as $D_{tr}  = \{(x, y) | x \in X, \ y \in Y_{tr} \}$ and $D_{tst} = \{(x, y) | x \in X, \ y \in Y_{tst} \}$ respectively.

The training set $D_{tr}$, and testing set $D_{tst}$ of an NIDS dataset can be represented as follows:

\begin{equation}
                D_{tr}  = \{(x, y) | x \in X_{tr}, \ y \in Y_{tr} \}
\end{equation}

\begin{equation}
                D_{tst} = \{(x, y) | x \in X_{tst}, \ y \in Y_{tst} \}
\end{equation}

\begin{equation*}
                where\ \  X_{tr} \subset X,\ \  X_{tst} \subset X 
\end{equation*}

\noindent here, $x$ represents a data sample (flow) chosen from the set of training  $X_{tr}$, test $X_{tst}$ data, and $X$ represents all the data samples. 
$y$ represents the corresponding labels and $Y_{tr}$ represents the set of class labels observed in the training phase, and
$Y_{tst}$ represents the set of class labels used in the test phase. 
In traditional Machine Learning, we have $Y_{tr} = Y_{tst}$, i.e., the set of classes observed during training is identical to the set of classes encountered by the model during testing.

% \begin{equation}
%                 T_{r} = T_{st} = \{(x, y) | x \in X, \ y \in \{b,\ a_1,\ a_2,\ ...,\ a_n\} \}
% \end{equation}
% \noindent  where $x$ is the data sample and $y$ is its respective class where $a_i, i\in\{1,...,n\}$ indicates the $i$th \textit{attack} class and $b$ represents the \textit{benign} class.

%%ZSL

The traditional ML setup has been commonly used in the ML-based NIDS evaluation process, proving to be effective in measuring the detection rate of known attack groups used in the training set. 
However, obtaining data samples for each attack class is very challenging for several different reasons. For instance, zero-day attacks have emerged repeatedly over the past few decades and present a serious risk to the  organisation of computer networks.
A zero-day attack can be a new kind or a modified threat that has not been seen or available earlier. 
The sole purpose of its design is to infiltrate and disrupt network communications \cite{bilge2012before}. Therefore, the traditional ML evaluation setup removes the conclusion that ML-based NIDSs are effective in the detection of zero-day attack scenarios, due to their unavailability at the time of training. 
ZSL techniques have been adopted to address such shortcomings in the evaluation of systems required to detect a larger set of classes than the one used in training. Unlike the traditional ML methods that focus on evaluating the generalization of the model to new data samples derived from pre-observed classes, in ZSL the objective is to improve the detection of unseen classes.

ZSL is essentially done in two stages \cite{xian2017zero}; an attribute learning stage where distinguishing knowledge, also known as auxiliary information, is captured. Followed by an inference stage where the learnt semantics are utilised to categorise data samples that belong to a new set of classes. ZSL is a promising approach to leverage supervised learning for the detection of unavailable training data samples. 
ZSL was principally developed to overcome the issue where none of the training samples is available. 
This approach overcomes the limitation of evaluating ML-based NIDS in the detection of zero-day attacks. 
As the collection and labelling of training data samples of zero-day attacks remain an impossible task simply due to their absence at the time of ML-based NIDS model development and training phases. 
Overall, ZSL overcomes the necessity of collecting training data samples of all the attack classes that the model will observe post-deployment including zero-day attacks.

%OUR ZSL
\begin{figure}[!b]
  \centering
  \includegraphics[width=13cm, height=5cm]{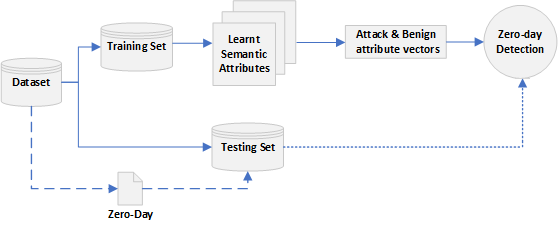}  
  \caption{Proposed Methodology}
  \label{architecture}
\end{figure}

In this paper, we propose a ZSL-based methodology, illustrated in Figure \ref{architecture}, to evaluate ML-based NIDSs in the detection of zero-day attacks. In the attribute learning stage, the model captures the semantic attributes of the attack behaviour using a set of known attacks and benign data samples. The attributes hold the distinguishing vectors between attack and benign network traffic. In the inference stage, the learnt knowledge is utilised to reconstruct the relationship between known attacks and the zero-day attack to classify the unseen zero-day attack group as malicious. Three main data concepts exist as part of the proposed methodology; 1) Known attacks- these are precedent attacks for which labelled data samples are available during training. 2) Zero-day attacks- these are unknown attacks that will emerge post-deployment for which labelled data samples are unavailable during training. 3) Semantic attributes- the distinguishing information that the ML model will learn from the known attacks to detect the zero-day attacks.

The proposed methodology assumes that at the testing stage, the model is evaluated using zero-shot samples derived from attack classes that were not available during the training stage. The model is required to detect the unseen class as malicious by associating the distinguishing information learnt from the observed and unobserved classes.

% \begin{equation}
%     % \begin{cases}
%         Y_{tr} = \{b,\ a_1,\ a_2,\ ..., a_n\}\setminus \{a_z\} \ \ \ \ for\ \ \ z \in \{1,...,n\}
% % \\
% % \\
% \end{equation}

% \begin{equation}
%     % \begin{cases}
%         Y_{tst} = \{b,\ a_1,\ a_2,\ ..., a_n\}
% % \\
% % \\
% \end{equation}

% Given a NIDS dataset, we define the set of traffic classes observed during training $Y_{tr}$ and the set observed during testing $Y_{tst}$ as follows:

\begin{figure}
  \centering
  \includegraphics[width=6cm, height=6cm]{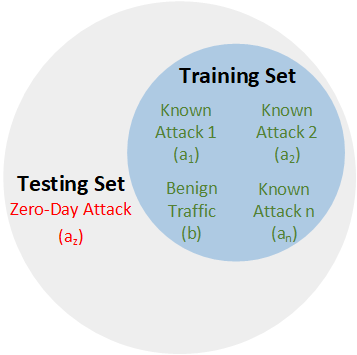}  
  \caption{Venn Diagram of the  training ($ D^z_{tr}$) and test ($ D^z_{tst}$) sets }
  \label{venn}
\end{figure}

% For the test of the proposed methodology we use the publicly available NIDS datasets. However, since we need the unseen attacks for the evaluation stage, we remove an attack class from the training set in each round. This procedure mathematically is stated below.
%
Given a NIDS dataset, we can define a ZSL training set $D^z_{tr}$ for an attack classes $z$ as follows:

\begin{equation}
    % \begin{cases}
        D^z_{tr} = \{(x, y) | x \in X_{tr}, \ y \in Y^z_{tr}=\{b,\ a_1,\ a_2,\ ..., a_n\}\setminus \{a_z\} \}\ \ \ \ for\ \ \ z \in \{1,...,n\}
% \\
% \\
\end{equation}

\begin{equation}
    % \begin{cases}
        D_{tst} = \{(x, y) | x \in X_{tst}, \ y \in Y_{tst}=\{b,\ a_1,\ a_2,\ ..., a_n\}\}
% \\
% \\
\end{equation}

\begin{equation*}
                where\ \  X_{tr} \subset X,\ \  X_{tst} \subset X 
\end{equation*}

\noindent the set of training classes $Y^z_{tr}$ consists of  benign traffic $b$, and $n$ attack classes $a_1, ..., a_n$, but importantly, minus the attack class $a_z$.
In contrast, the test dataset $D_{tst}$ always consists of samples of all classes, i.e., without the removal of any attack class.
By excluding an attack class $z$ from the training phase, we are essentially simulating a zero-day attack scenario, as the ML model has not been trained on the respective attack class and a new attack class has emerged post the training phase. 
The concept of our ZSL evaluation scenario is also illustrated in Figure \ref{venn}.

% \begin{equation}
%     T^z_{st} = \{(x, y) | x \in X, \ y \in \{b,\ a_1,\ a_2,\ ..., a_n\} \}
%     % \end{cases}
% \end{equation}

% \begin{equation}
%     % \begin{cases}
%         T^z_{r} = \{(x, y) | x \in X, \ y \in \{b,\ a_1,\ a_2,\ ..., a_n\}\setminus \{a_z\} \}\ \ \ \ for\ \ \ z \in \{1,...,n\}
% % \\
% % \\
% \end{equation}
% \begin{equation}
%     T^z_{st} = \{(x, y) | x \in X, \ y \in \{b,\ a_1,\ a_2,\ ..., a_n\} \}
%     % \end{cases}
% \end{equation}
% \noindent in which $a_z$ is the zero-day attack class, $\mathit{T^z_{r}}$ indicates the ZSL training set corresponding to the attack class, and $T^z_{st}$ represents the ZSL test set that includes the zero-day attack class.

%  During the training phase, a single attack class is removed from the dataset, this is referred to as the zero-day attack. The model is trained using the training set, which does not include the zero-day attack as indicated by Figure \ref{venn}. 
 
% \textcolor{red}{
The purpose of this stage is for the model to map the network data feature space to the relevant attribute space defining the attack behaviour. The learnt semantics can be used to distinguish a zero-day attack from benign traffic. Assuming $S$ is the set of
% } \textcolor{blue}{samples of} \textcolor{red}{ 
known attack classes that can be used to train an ML model, each known attack data sample is denoted by $x$, their respective labels denoted by $y$, and the semantic information of attack behaviour is denoted by $h$. Therefore, $x$ and $y$ can be values of one of the known attack samples and classes, respectively. This can be represented using the following notation, 
% }

\begin{equation}
        S = \{(x, y, h) | x \in X_{tr}, \ y \in \{a_1,\ a_2,\ ..., a_n\}\setminus \{a_z\},  h \in H \}\ \ \ \ for\ \ \ z \in \{1,...,n\}
\end{equation}

\noindent where $H$ is the set of learnt attributes that can be used to predict zero-day attacks.

During the testing or inference phase, the zero-day attack traffic class $a_z$ is added back to the test set, in order to measure the zero-day detection accuracy. For this purpose, we have defined a new evaluation metric, which is discussed in  Section \ref{zdr}. This follows the generalised ZSL setting where the test samples may belong to the seen (known attacks and benign traffic) or unseen (zero-day attack) data samples \cite{felix2019generalised}. This has proven to be a more practical scenario than the conventional ZSL setting, where the test set only includes samples from the unseen class, which is difficult to guarantee from a network security perspective. The main goal in this setting is to reconstruct the knowledge learnt from the known attacks so that the model can detect the zero-day attack as malicious. This is accomplished by associating the relationships between known attacks and zero-day attacks. For each of the available attack classes in the dataset, we simulate the zero-day attack scenario by removing this class from the corresponding training set and consider the ability of the model trained on all the other attack classes to detect the unseen attack class.

The proposed zero-day detection evaluation methodology aims to evaluate ML-based NIDSs in regards to their ability to generalise and detect new and unseen attack classes post-deployment, which is a very realistic and relevant scenario.   
Information gained from our evaluation methodology can be used to further optimise the machine learning model's hyper parameters and architecture, in order to enhance the generalisability to new attack types. Furthermore, network data feature selection experiments can be conducted to identify key features required to predict the behavioural attack patterns to detect zero-day attacks. The reliable detection of zero-day attacks is essentially one of the key limitations faced by existing signature-based NIDS, due to the inability of identifying the complete IOCs related to a future attack class. Hence, the practical motivation for organisations switching to an ML-based NIDS is the detection and prevention of zero-day attacks, which is the focus of our proposed method.

%https://learnopencv.com/zero-shot-learning-an-introduction/

%%%%%%%%%%%%%here
\section{Experimental Setup}
\label{method}
The evaluation of ML-based NIDS capability to detect zero-day attacks using a set of attributes learnt from known attacks is crucial. As the performance motivates the usage and development of ML in the detection of zero-day attacks. In this paper, two commonly used ML models have been used in the design of ML-based NIDSs, Random Forest (RF) \cite{breiman2001random} and Multi-Layer Perceptron (MLP) \cite{hinton1990connectionist}. They have been deemed to achieve reliable performance and typically achieve a high detection accuracy. The RF classifier is designed using randomised 50 decision trees classifiers in the forest, each following the Gini impurity function \cite{breiman1996some} to measure the quality of a split. The MLP neural network model is structured with 100 neurons in two hidden layers, each performing the Rectified Linear Unit (ReLU) \cite{agarap2018deep} activation function. The stochastic gradient-based optimiser is utilised for the model's weight and parameters optimisation. In the inference stage, a five-fold cross-validation method is adopted to calculate the mean results. 

\subsection{Datasets}
In this paper, two NIDS datasets are used to evaluate the ML models following the proposed methodology, i.e., UNSW-NB15 \cite{moustafa2015unsw}, and NF-UNSW-NB15-v2 \cite{sarhan2021towards}. The datasets are synthetic which were created via virtual network testbeds representing modern network structures. In designing such datasets, certain attack scenarios are conducted and the corresponding network traffic is captured and labelled with the respective attack type. In addition, normal network traffic is generated representing benign traffic is captured and labelled accordingly. Both the malicious and non-malicious traffic is captured in the native packet capture (pcap) format, and certain data features are extracted to represent explicit information regarding the data flow. The chosen datasets include a variety of modern network attacks such that each can be used to simulate the incoming of a zero-day attack. Such datasets have been widely used in the literature as they do not present the limitations faced by the collection and labelling of real-world production networks. 

\begin{itemize}
    \item \textbf{UNSW-NB15 \cite{moustafa2015unsw}}- A well-known and widely used NIDS dataset in the research community released in 2015 by the Cyber Range Lab of the Australian Center for Cyber Security (ACCS). The synthetic dataset is designed using the IXIA Perfect Storm tool to generate benign network activities and premeditated attack scenarios. The dataset contains  49 features extracted by Argus and Bro-IDS tools and twelve additional SQL algorithms. The dataset consists of 2,218,761 (87.35\%) benign samples and 321,283 (12.65\%) attack on, that is, 2,540,044 network data samples in total.
    
    \item \textbf{NF-UNSW-NB15-v2 \cite{sarhan2021towards}}- A recently released dataset in 2021 by the University of Queensland. The dataset is generated by extracting 43 NetFlow-based features from the pcap files of the UNSW-NB15 dataset. The nprobe feature extraction tool is utilised to extract the network data flows from the pcap files. The total number of data flows are 2,390,275 out of which 95,053 (3.98\%) are attack samples and 2,295,222 (96.02\%) are benign.
\end{itemize}

The complete set of network data samples in each dataset is used in this paper. Initially, the flow identifiers such as sample id, source/destination IPs, source/destination ports, and timestamps are dropped to avoid learning bias towards the attacking and victim endpoints. This is required as distinct nodes in the testbed have been used to launch attack scenarios targeting certain network ports. Moreover, all categorical-based features are converted to numerical-based values using the label encoding technique where each label is assigned a unique integer. Once a full numerical dataset is obtained to accommodate for efficient experiments, the min-max scaler technique is applied to normalise all values between 0 and 1. This is necessary to avoid the learning model from assigning higher weights to features holding larger numerical values. This completes the preprocessing stage where the data is ready for efficient ML training and testing stages following the proposed setup.  

\subsection{Zero-day Detection Rate}
\label{zdr}

\renewcommand{\arraystretch}{1.5}
\begin{table}[b!]\footnotesize	
\centering
\caption{Evaluation Metrics}
\begin{tabular}{|>{\centering\arraybackslash}m{4cm}|>{\centering\arraybackslash}m{5.5cm} |>{\centering\arraybackslash}m{3cm} |}

\hline
\rowcolor{lightgray}
\textbf{Metric}            & \textbf{Definition}           & \textbf{Equation}                               \\ \hline
Accuracy                   & The percentage of correctly classified samples in the test set.              &\normalsize $\frac{TP+TN}{TP+FP+TN+FN} \times 100$\\ \hline
Detection Rate (DR)        & The percentage of correctly classified total attack samples in the test set. &\normalsize $\frac{TP}{TP+FN} \times 100 $\\ \hline

False Alarm Rate (FAR)     & The percentage of incorrectly classified benign samples in the test set. &  \normalsize $\frac{FP}{FP+TN} \times 100$   \\ \hline

Area Under the Curve (AUC) & The area underneath the DR and FAR plot curve in the test set.        & N/A       \\ \hline
F1 Score                   & The harmonic mean of the model's precision and DR.  & \normalsize $2 \times \frac{DR\;\times \;Precision}{DR\; +\; Precision}$         \\ \hline 
\textbf{Zero-day Detection Rate ($Z$-$DR_z$)} & \textbf{The percentage of correctly classified zero-day attack samples in the test set.} & \normalsize{$\frac{\textbf{TP}_{\textbf{a}_\textbf{z}}}{\textbf{TP}_{\textbf{a}_\textbf{z}}+\textbf{FN}_{\textbf{a}_\textbf{z}}}\times \textbf{100}$}
\\ \hline 
\end{tabular}%
% }
\label{eme}
\end{table}
\renewcommand{\arraystretch}{1}

% \begin{figure}[b]
%   \centering
%   \includegraphics[width=6cm]{Multiclass_CF.pdf}  
%   \caption{Multiclass Confusion Matrix}
%   \label{fig: MC-CF}
% \end{figure}

% The standard evaluation metrics listed in Table \ref{eme} are collected to evaluate the ML models' performance, which are based on the number True Positives (TP), False Positives (FP), True Negatives (TN), and False Negatives (FN).  

We will use the common classification performance metrics of  Accuracy, Detection Rate (DR), False Alarm Rate (FAR), Area under the (ROC) Curve (AUC), and F1 Score for our evaluation. 
These metrics are defined based on the True Positives (TP), False Positives (FP), True Negatives (TN), and False Negatives (FN) number, as shown in Table \ref{eme}.
In our evaluation scenarios, these metrics are calculated based on a binary classification case, where the classifier distinguishes between benign and attack traffic, and is hence equivalent to the micro-average in the multiclass classification scenario. 

In addition to these standard metrics, we define a new evaluation metric called Zero-Day Detection Rate ($Z$-$DR_z$), also shown in Table \ref{eme}, which is defined as the specific detection rate of the zero-day attack class $a_z$, which was excluded from the training data set (Equation \ref{zz}).

%-------------------------------->>>>>>>
%-------------------------------------------->>>>>>>>    WORK IN PROGRESS  ---->

%%%%%%%%%%
% In addition, as part of the proposed methodology, a new evaluation metric called Z-DR is defined in Equation \ref{zz} to measure the detection rate of each zero-day attack scenario. $ZDR_z$ refers to the specific detection rate of attacks of type $a_z$, which were not included in the corresponding training set $D^z_{tr}$.  

\begin{equation}
    Z\mbox{-}DR_z = \frac{TP_{a_z}}{TP_{a_z}+FN_{a_z}}\times 100
    \label{zz}
\end{equation}

Here, $TP_{a_z}$ and $FN_{a_z}$ are the number of True Positives and False Negatives calculated specifically for the samples of the zero-day attack class $a_z$. 
The new metric measures how well the ML model can detect zero-day attacks of class $a_z$, i.e. based solely on training information provided by the other attack classes. 
It therefore indicates the ability of the ML model to generalise the information learnt from the other attack classes to the new, unseen (zero-day) class.

% In addition, as part of the proposed methodology, a new evaluation metric called Z-DR is defined in Equation \ref{zz} to measure the detection rate of each zero-day attack scenario. 

% \begin{equation}
%     \frac{TP_{a_z}}{TP_{a_z}+FN_{a_z}}\times 100
%     \label{zz}
% \end{equation}

%%%%%%%%%%%%%%%%%%%%%%%%%%%%%%%%%%%%%%%%%%%%%%%%%%%%%%%%%%%%
\section{Evaluation}
\label{results}

In this section, two ML models, Multilayer Perceptron (MLP)  and Random Forest (RF), have been utilised to evaluate the detection of zero-day attacks using our proposed ZSL evaluation scenario. Two synthetic NIDS datasets (UNSW-NB15 and NF-UNSW-NB15-v2) have been used in the experiments. Each available attack in the datasets is considered to simulate a zero-day attack incident. The models are evaluated based on the Z-DR, as well as the overall detection accuracy of the test set that includes known attacks, zero-day attack, and benign data samples. This represents a generalised ZSL setup where the test set includes both known and unknown data samples, which is appropriate for ML-based NIDS evaluation.

\subsection{Results}\label{res}
The goal of the experiments is to evaluate the performance of ML-based NIDSs in a realistic network environment where attacks not used in the training of the model are likely to be observed post-deployment. Tables \ref{1} to \ref{4} display the complete set of results collected. Each table represents a unique combination of an ML model and a dataset. In each table, the first column lists the attacks used to simulate a zero-day attack incident. The second column displays the corresponding Z-DR value, and the rest of the columns present the remaining evaluation metrics collected over the complete test set, which include the zero-day attack, known attacks and benign data samples.

\begin{table}[!b]\scriptsize
\centering
\caption{MLP UNSW-NB15}
\begin{tabular}{|l|l|l|l|l|l|l|}
\hline
\textbf{Zero-day Attack} & \textbf{Z-DR} & \textbf{Accuracy} & \textbf{F1 Score} & \textbf{FAR} & \textbf{DR} & \textbf{AUC} \\ \hline
\textbf{Exploits}             & \textbf{90.31}     & 98.73             & 0.92              & 0.47         & 89.09       & 0.94                               \\ \hline
\textbf{Fuzzers}              & \textbf{20.10}     & 96.94             & 0.74              & 0.15         & 59.16       & 0.80                               \\ \hline
\textbf{Generic}              & \textbf{95.90}     & 98.93             & 0.93              & 0.36         & 90.09       & 0.95                                \\ \hline
\textbf{Reconnaissance}       & \textbf{91.82}     & 98.93             & 0.91              & 0.48         & 90.17       & 0.95                               \\ \hline
\textbf{DoS}                  & \textbf{92.80}     & 99.0             & 0.91              & 0.35         & 87.70       & 0.94                               \\ \hline
\textbf{Analysis}             & \textbf{84.35}     & 99.06             & 0.91              & 0.53         & 91.36       & 0.95                              \\ \hline
\textbf{Backdoor}             & \textbf{99.04}     & 99.03             & 0.91              & 0.60         & 92.10       & 0.96                               \\ \hline
\textbf{Shellcode}            & \textbf{97.15}     & 99.08             & 0.91              & 0.48         & 90.67       & 0.95                               \\ \hline
\textbf{Worms}                & \textbf{98.25}     & 99.06             & 0.90              & 0.51         & 90.60       & 0.95                              \\ \hline
\end{tabular}
\label{1}
\end{table}

In Table \ref{1} and \ref{2}, the performance of the MLP and RF classifiers, when evaluated using the UNSW-NB15 dataset, are presented. During the simulation of zero-day attacks, the Exploits, Reconnaissance, and DoS attacks are detected at a rate of around 90\% using the MLP classifier. The RF classifier is more effective in the detection of Exploits and DoS attacks. Both the MLP and RF models detect 20\% and 15\% of the fuzzer attack data samples, respectively. This kind of attack presents severe risks to organisations protected by such vulnerable ML-based NIDSs in the scenario of a launched zero-day attack similar to Fuzzers. The MLP model is superior to RF in the detection of Generic and Shellcode attack types, achieving a high detection rate of 96\% and 97\% compared to 59\% and 91\%, respectively. The Analysis attack type is deemed to be complex in its detection as a zero-day attack where the MLP model achieved an 84\% and the RF model detected 81\%. Other attack types, such as Backdoor and Worms, were almost fully detected by both ML models when observed as zero-day attacks.

\begin{table}[!t]\scriptsize
\centering
\caption{RF UNSW-NB15}
\begin{tabular}{|l|l|l|l|l|l|l|}
\hline
\textbf{Zero-day Attack} & \textbf{Z-DR} & \textbf{Accuracy} & \textbf{F1 Score} & \textbf{FAR} & \textbf{DR} & \textbf{AUC} \\ \hline
\textbf{Exploits}        & \textbf{94.43}              & 99.07             & 0.94              & 0.33         & 91.95       & 0.96         \\ \hline
\textbf{Fuzzers}         & \textbf{14.77}              & 96.92             & 0.73              & 0.06         & 57.58       & 0.79         \\ \hline
\textbf{Generic}         & \textbf{59.06}              & 97.64             & 0.82              & 0.38         & 73.19       & 0.86         \\ \hline
\textbf{Reconnaissance}  & \textbf{89.08}              & 99.05             & 0.93              & 0.36         & 90.26       & 0.95         \\ \hline
\textbf{DoS}             & \textbf{96.89}              & 99.25             & 0.93              & 0.36         & 92.52       & 0.96        \\ \hline
\textbf{Analysis}        & \textbf{81.37}              & 99.22             & 0.92              & 0.36         & 91.37       & 0.96         \\ \hline
\textbf{Backdoor}        & \textbf{99.60}              & 99.28             & 0.93              & 0.37         & 92.58       & 0.96         \\ \hline
\textbf{Shellcode}       & \textbf{90.80}              & 99.25             & 0.92              & 0.35        & 91.59       & 0.96         \\ \hline
\textbf{Worms}           & \textbf{100.00}             & 99.28             & 0.93              & 0.37         & 92.24       & 0.96         \\ \hline
\end{tabular}
\label{2}
\end{table}

The performance of both ML models were depended on the complexity of the incoming zero-day attacks. The models were successful in the detection of the 95\% or more of attacks such as Generic, DoS, Backdoor, Shellcode, and Worms. However, Exploits, Reconnaissance, and Analysis have shown to be harder to detect, with both models achieving around 90\% detection rates. However, in the likely scenario of the models observing attacks related to the Fuzzers attack group as a zero-day attack, ML-based NIDSs would be extremely vulnerable as more than 80\% of their data samples were undetected and classified as benign samples. Further Analysis is required to investigate the complexity of the Fuzzers attack group that causes the undetected zero-day attack samples by the ML models. Overall, the MLP classifier achieved an average of 85.5\% detection rates across the zero-day attacks. The RF classifier was slightly inferior with an average detection rate of 80.67\%.

\begin{figure*}[!b]
\begin{subfigure}{.5\textwidth}
  \centering
  \includegraphics[width=8cm, height=4cm]{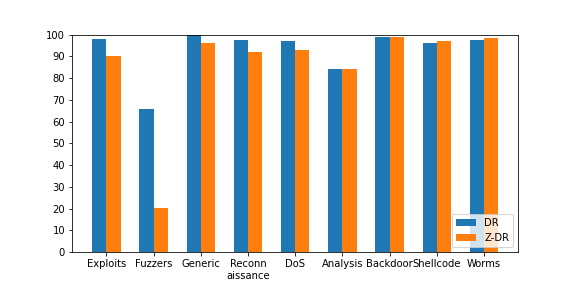}  
  \caption{MLP}
  \label{11}
\end{subfigure}
\hfill
\begin{subfigure}{.5\textwidth}
  \centering
  \includegraphics[width=8cm, height=4cm]{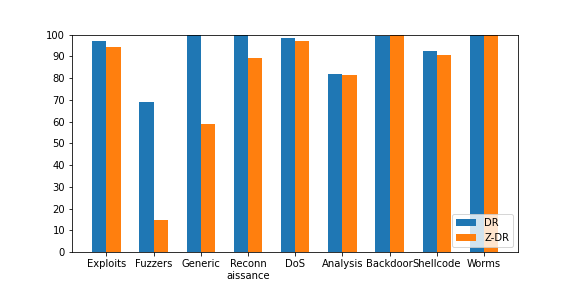}  
  \caption{RF}
  \label{22}
\end{subfigure}
\caption{Comparison between DR vs Z-DR of attacks in UNSW-NB15}
\label{unsw}
\end{figure*}

In Figure \ref{unsw}, the detection rate of each attack group in the UNSW-NB15 dataset is measured in the (traditional) known attack and zero-day attack scenarios. In the known attack scenario, the model has observed the attack in the training set. In the zero-day attack scenario, the attack is not available in the training set for the model to observe. Figures \ref{11} and {22} represent the performance using the MLP and RF models, respectively. The drop of detection rate is highly notable in certain attack types such as Fuzzers and Reconnaissance. The DR value dropped by around 70\% and 10\% respectively for the two ML models. Furthermore, there are distinct differences in the performance of the two models. The MLP model was more successful in the detection of zero-day Generic attacks at a detection rate of 95.90\% compared to 59.06\% achieved by RF. Both models achieved a 100\% detection rate when the attack class was observed in the training set. The RF classifier has been slightly more efficient in the detection of the Exploits and DoS zero-day attack groups.

\begin{table}[h]\scriptsize
\centering
\caption{MLP NF-UNSW-NB15-v2}
\begin{tabular}{|l|l|l|l|l|l|l|}
\hline
\textbf{Zero-day Attack} & \textbf{Z-DR} & \textbf{Accuracy} & \textbf{F1 Score} & \textbf{FAR} & \textbf{DR} & \textbf{AUC} \\ \hline
\textbf{Exploits}          & \textbf{81.47}     & 98.89             & 0.92              & 0.29         & 87.74       & 0.94                               \\ \hline
\textbf{Fuzzers}              & \textbf{76.19}     & 98.96             & 0.91              & 0.19         & 85.72       & 0.93                               \\ \hline
\textbf{Generic}              & \textbf{99.57}     & 99.62             & 0.97              & 0.33         & 98.85       & 0.99                                \\ \hline
\textbf{Reconnaissance}       & \textbf{99.75}     & 99.60             & 0.96              & 0.30         & 97.89       & 0.99                               \\ \hline
\textbf{DoS}                  & \textbf{90.68}     & 99.55             & 0.95              & 0.31         & 96.53       & 0.98                               \\ \hline
\textbf{Analysis}             & \textbf{88.47}     & 99.60             & 0.95              & 0.34         & 98.23       & 0.99                              \\ \hline
\textbf{Backdoor}             & \textbf{97.28}     & 99.59             & 0.95              & 0.29         & 96.90       & 0.98                               \\ \hline
\textbf{Shellcode}            & \textbf{98.60}     & 99.63             & 0.96              & 0.30         & 97.94       & 0.99                               \\ \hline
\textbf{Worms}                & \textbf{100.00}     & 99.63             & 0.95              & 0.32         & 98.46       & 0.99                              \\ \hline
\end{tabular}
\label{3}
\end{table}

In Tables \ref{3} and \ref{4}, the ML models' zero-day attack detection performance is evaluated using NF-UNSW-NB15-v2, the NetFlow-based edition of the UNSW-NB15 dataset. The MLP model is superior to the RF model in the detection of Exploits and Fuzzers zero-day attack groups with a detection rate of 82\% and 76\% compared to 59\% and 51\%, respectively. The ML models did not successfully apply the learnt semantic attributes of the attack behaviour to be able to relate the Exploits and Fuzzers zero-day attacks as malicious traffic. Based on the results of our considered scenario, we can say that attacks such as Generic, Reconnaissance, Backdoor, Shellcode, and present a significantly lower cybersecurity risk to organisations protected by ML-based NIDS when they are observed for the first time as zero-day attacks. The utilised models correctly detected close to 100\% of their data samples as intrusive traffic. Moreover, DoS and Analysis attack groups were slightly harder to detect, as both the ML models detected around 90\% of their data samples.

\begin{table}[!b]\scriptsize
\centering
\caption{RF NF-UNSW-NB15-v2}
\begin{tabular}{|l|l|l|l|l|l|l|}
\hline
\textbf{Zero-day Attack} & \textbf{Z-DR} & \textbf{Accuracy} & \textbf{F1 Score} & \textbf{FAR} & \textbf{DR} & \textbf{AUC} \\ \hline
\textbf{Exploits}        & \textbf{59.28}              & 98.07             & 0.84              & 0.11         & 73.33       & 0.87         \\ \hline
\textbf{Fuzzers}         & \textbf{51.32}              & 98.38             & 0.85              & 0.10         & 74.61       & 0.87         \\ \hline
\textbf{Generic}         & \textbf{99.11}              & 99.75             & 0.98              & 0.15         & 98.05       & 0.99         \\ \hline
\textbf{Reconnaissance}  & \textbf{99.57}              & 99.77             & 0.98              & 0.15         & 98.15       & 0.99         \\ \hline
\textbf{DoS}             & \textbf{93.68}              & 99.71             & 0.97              & 0.15         & 96.85       & 0.98         \\ \hline
\textbf{Analysis}        & \textbf{87.95}              & 99.75             & 0.97              & 0.14         & 97.15       & 0.99         \\ \hline
\textbf{Backdoor}        & \textbf{99.49}              & 99.76             & 0.97              & 0.16         & 97.84       & 0.99         \\ \hline
\textbf{Shellcode}       & \textbf{95.94}              & 99.75             & 0.97              & 0.16         & 97.70       & 0.99         \\ \hline
\textbf{Worms}           & \textbf{100.00}             & 99.77             & 0.97              & 0.14         & 97.59       & 0.99         \\ \hline
\end{tabular}
\label{4}
\end{table}

Most of the attacks present in the NF-UNSW-NB15-v2 dataset were reliably detected using the considered two ML models in the considered zero-day attack scenario. The learning models were successful in utilising the learnt information from known attacks to detect the zero-day attack types. However, the Exploits and Fuzzers attack scenarios seem to be harder to detect, if the ML model did not observe them during training and encounter them as zero-day attacks. 
 Overall, the MLP and RF models average Z-DR values of 92.45\% and 87.37\%, respectively.The UNSW-NB15 and NF-UNSW-NB15-v2 datasets contain the same attack groups, and they differ only in their respective feature sets. The NetFlow-based feature set of NF-UNSW-NB15-v2 results in an increased detection rate by around 7\% for each of the two ML models. This demonstrates the advantage of using NetFlow-based features in the detection of zero-day attack scenarios.

\begin{figure*}[h!]
\begin{subfigure}{.5\textwidth}
  \centering
  \includegraphics[width=8cm, height=5cm]{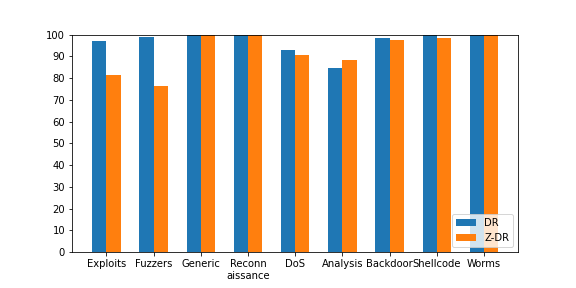}  
  \caption{MLP}
  \label{33}
\end{subfigure}
\hfill
\begin{subfigure}{.5\textwidth}
  \centering
  \includegraphics[width=8cm, height=5cm]{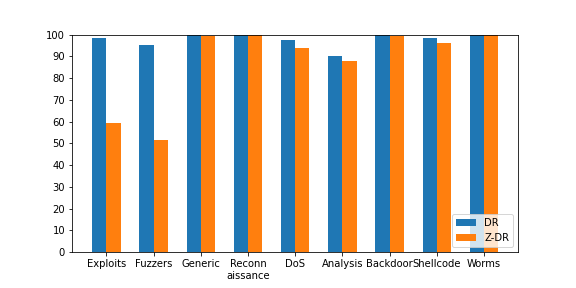}  
  \caption{RF}
  \label{44}
\end{subfigure}
\caption{Comparison between DR vs Z-DR of attacks in NF-UNSW-NB15-v2}
\label{nf}
\end{figure*}

In Figure \ref{nf}, the detection rate of each attack group in the NF-UNSW-NB15-v2 dataset is shown for both known attack and zero-day attack scenarios. Figures \ref{33} and \ref{44} display the performance of the MLP and RF models, respectively. In this dataset, a large drop in detection rates is noted for the Exploits and Fuzzers attack groups, with an average decrease of 28\% and 35\%, respectively for the two ML models in a zero-day attack scenario. For the rest of the attack groups, the ML models were successfully able to detect the attacks, however, the DoS and Analysis were slightly sophisticated in their detection even in a known attack scenario.

\subsection{Analysis}

To investigate the results provided in the previous subsection, especially the low Z-DRs of particular attacks, the feature distribution of each attack group is studied in both datasets. The main objective of this analysis is to find out any possible differences between several ZSL training and testing sets. As such, many statistical measures were explored that could identify the dissimilarities between the feature distributions. The \textit{Wasserstein Distance (WD)} metric, which is commonly used in the ML/AI community, has been successfully used in~\cite{l2021bench} for quantifying the feature distribution distances. The (first) Wasserstein Distance, also known as the \textit{Earth Mover’s distance}, 
is a distance function defined between two probability distributions $u$ and $v$ and is defined as follows \cite{Ramdas2017}:

\begin{equation}\label{wd}
W(u,v) = \inf_{\gamma \in \Gamma(u,v) } \int_{\mathbb{R} \times \mathbb{R} } |x-y| d\gamma(x,y)
\end{equation}
\\
\noindent Here, $\Gamma(u,v)$ is the set of (probability) distributions on $\mathbb{R} \times \mathbb{R}$ where $u$ and $v$ are its first and second factor marginals. $\gamma(x,y)$ can be interpreted as a  transport plan/function that gives the amount of mass to move from each $x$ to $y$ to transport $u$ to $v$, subject to the following constraints:

\begin{equation}
\begin{cases}
   \displaystyle \int{\gamma(x,y)dy} = u(x) \\
    \\
    \displaystyle \int{\gamma(x,y)dx} = v(y) 
\end{cases}
\end{equation}

\noindent this indicates that for an infinitesimal region around $x$, the total mass moved out must be equal to ${u(x)dx}$ and similarly, for an infinitesimal region around $y$ the total mass moved in must be equal to $v(y)dy$.

Using WD as the comparison metric, a set of experiments were conducted to investigate the differences in the feature distributions of 9 different zero-day scenarios (one per attack class in each dataset). After selecting the training ($D^z_{tr}$) and testing ($D^z_{tst}$) sets, for each data feature (except the flow identifier features that were removed in the pre-processing stage), the feature distributions between the sets were compared using the WD metric, i.e. $W(D^z_{tr}, D^z_{tst})$ in the form of Equation~\ref{wd} notation. The method is performed by measuring the WD between the set of known attacks and the set including the zero-day attack. This is the same setup as performed in the zero-shot learning and inference phases. Hence, in each zero-day attack scenario, a WD value corresponding to each feature of the dataset was obtained. A higher WD value for a feature indicates a more distinctive distribution between the training ($D^z_{tr}$) and testing ($D^z_{tst}$) sets of the corresponding zero-day attack.

\begin{figure*}[h!]
\begin{subfigure}{.5\textwidth}
  \centering
  \includegraphics[width=7.5cm, height=5cm]{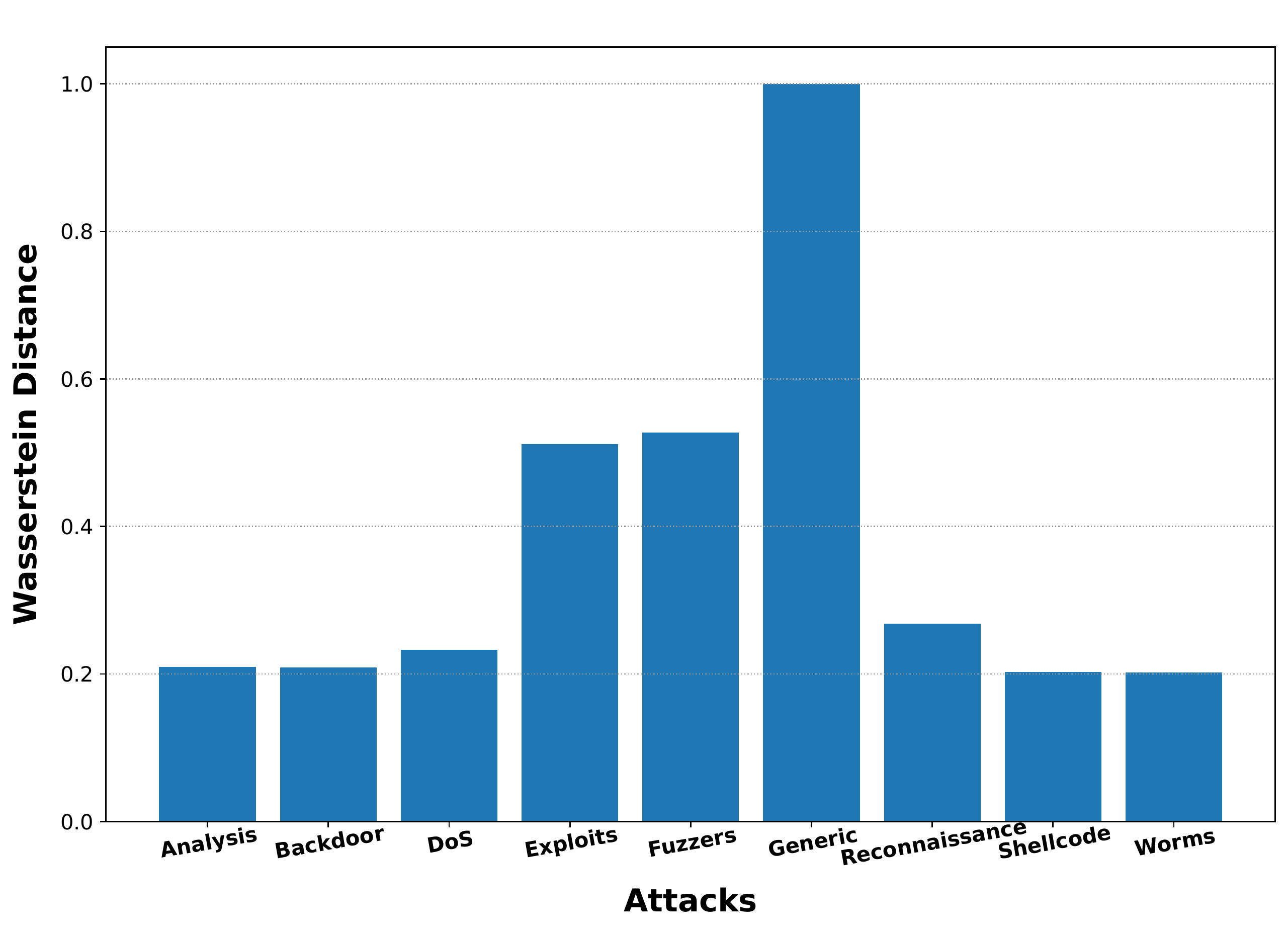}
  \caption{UNSW-NB15}
  \label{55}
\end{subfigure}
\hfill
\begin{subfigure}{.5\textwidth}
  \centering
  \includegraphics[width=7.5cm, height=5cm]{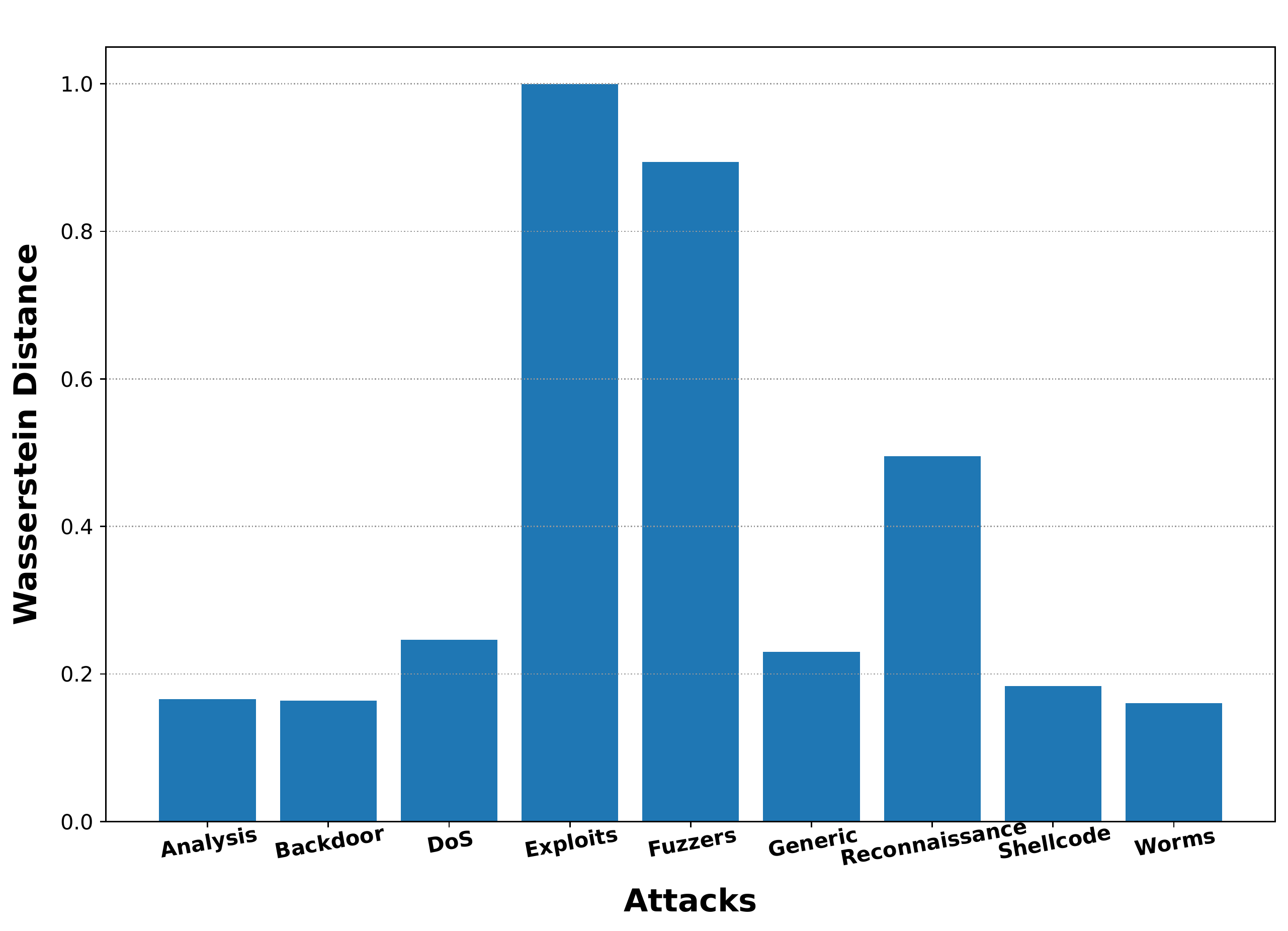}  
  \caption{NF-UNSW-NB15-v2}
  \label{66}
\end{subfigure}
\caption{Wasserstein distances of feature distributions between the train and test sets, $W(D^z_{tr}, D^z_{tst})$, of each zero-day attack, averaged over all features, a) UNSW-NB15 dataset, and b) NF-UNSW-NB15-v2 dataset}
\label{ws}
\end{figure*}

Figures \ref{55} and \ref{66}, show the WD value of each zero-day attack scenario, averaged over all features in each experiment for the UNSW-NB15 and NF-UNSW-NB15-v2 datasets respectively. As can be observed, most attacks have a low WD value of around 0.2, which indicates the overall feature distributions are similar between the training and testing sets in the case of these zero-day attack scenarios. This shows that these attacks are similar in their statistical feature distributions to the rest  of the attacks. 
Due to the similarity in the attack types, it is expected to see a higher zero-day detection performance. This mostly includes the Analysis, Backdoor, DoS, Reconnaissance, Shellcode, and Worms attacks. Considering Tables~\ref{1}, \ref{2}, \ref{3} and \ref{4} and the Z-DR values of these attacks, we see that these attacks are detected with a high detection rate in a zero-day attack scenario. Our results show that there is only a minor degradation in their Z-DR values compared to their normal (non-zero-day) detection rate, using the same ML model.

The correspondence between the WD values and the Z-DR performance indicates that the similarity of the feature distributions, between the zero-day training and testing sets makes it possible for the model to learn useful patterns about the zero-day attack from the features of the known attack classes. Hence, as long as the feature distributions are not significantly different between the training and test sets in a zero-day scenario, the ML-based NIDS is expected to classify the unseen attack (classes) with a classification performance similar to if it had been seen in the training stage. Such attacks represent a relatively lower risk to organisations adopting an ML-based NIDS due to the similar statistical patterns that ML models can utilise in their detection.

These results are further confirmed by the results related to the three remaining attack groups, i.e., Exploits, Fuzzers, and Generic. As per the UNSW-NB15 dataset, the high WD value for Fuzzers and Exploits with an averaged value of 0.5, and the Generic class with an averaged WD value of 1.0 indicates that the overall feature distributions are much more distinctive between the zero-day training and testing sets. Accordingly, we can expect to see a worse degradation of their Z-DR compared to known attack scenarios, which is confirmed by the results shown in Table~\ref{1} and \ref{2}. Similarly, for the NF-UNSW-NB15 dataset, the Fuzzers and Exploits stand out with very high average WD values of 0.9 and 1.0 respectively. Their corresponding Z-DR values are much lower than for other attack classes, as per Tables~\ref{3} and \ref{4}. The findings are consistent due to the distinguishing attributes of such attack groups that were not learnt via the seen classes. 

Overall, the WD function has identified several attack groups with a unique malicious pattern compared to the remainder of the attacks. Therefore, from an ML perspective, their detection as zero-day attacks using an ML-based NIDS will be challenging. This matches the results in this paper, as there is a significant difference between their Z-DR and the seen detection rate. Further studies are required to improve ML-based NIDSs in the detection of unique attack behaviour related to sophisticated attacks. 

While not perfect, our Analysis using the Wasserstein (WD) distance between feature distributions of different attack classes provides a solid explanation of the results presented in sub-section (Sub-Section~\ref{res}), and is consistent with the main findings of this paper.

\section{Conclusion}
ZSL is a key technique used to improve an ML model's ability to classify data samples derived from classes that have not been accessible during the training stage. In this paper, a ZSL-based methodology has been proposed to evaluate the performance of an ML-based NIDS in the detection of unseen, also known as zero-day, attacks. In the attribute learning stage, the model learns the distinguishing attributes of attack traffic using a set of known attacks. This is accomplished by mapping relationships between the network data feature space to the semantic space. In the inference stage, the model is required to associate the learnt attributes with the attack behaviour to detect a zero-day attack that was not observed during the training stage. Using our proposed methodology, two well-known ML models have been used to evaluate their ability to detect each attack present in the UNSW-NB15 and NF-UNSW-NB15-v2 datasets as a zero-day attack. The results show that an ML-based NIDS is, somewhat surprisingly, capable of securing organisations against most zero-day attacks to a significant extent. 

However, certain attack groups, such as Fuzzers, Exploits, and Analysis
have shown a much lower Zero-day detection rate, and the ML model trained on all other attack types was unable to generalise to these types of attacks. Our Analysis using the Wasserstein Distance (WD) distance metric of the feature distributions is able to explain this to some extent. We observed that the WD metric of the different attack types correlates strongly with their Zero-day detection rate.  The statistical feature distributions of these attacks are too different from the other attack types observed during the training stage, and prevent the model to generalise and detect them in the test phase of the experiment.
The ability to generalise and detect new, unseen types of attacks is an important feature of ML-based NIDSs, in is critical for increased practical deployment in production networks.
This important issue has so far attracted only relatively limited attention in the research literature. We hope that the work presented in this paper provides a basis and motivation for further research into this important aspect.

\bibliography{main.bib}

\end{document}